\pdfoutput=1

\documentclass[11pt]{article}

\usepackage{emnlp2021}

\usepackage{times}
\usepackage{latexsym}

\usepackage[T1]{fontenc}

\usepackage[utf8]{inputenc}

\usepackage{microtype}

\usepackage{amssymb}
\usepackage{amsmath}
\usepackage{mathrsfs}
\usepackage{graphicx}
\usepackage{booktabs}
\usepackage{url}
\usepackage{array}
\usepackage{multirow}
\usepackage{subfigure}
\usepackage{tablefootnote}
\usepackage{threeparttable}
\usepackage{hyperref}

\usepackage{enumitem}
\setenumerate[1]{itemsep=0pt,partopsep=0pt,parsep=\parskip,topsep=5pt}
\setitemize[1]{itemsep=0pt,partopsep=0pt,parsep=\parskip,topsep=5pt}
\setdescription{itemsep=0pt,partopsep=0pt,parsep=\parskip,topsep=5pt}

\title{Contextualize Knowledge Bases with Transformer for End-to-end Task-Oriented Dialogue Systems}

\author{Yanjie Gou\textsuperscript{\rm 1}, 
Yinjie Lei\textsuperscript{\rm 1}\thanks{\ \  Corresponding author}, Lingqiao Liu\textsuperscript{\rm 2}, Yong Dai\textsuperscript{\rm 3}, Chunxu Shen\textsuperscript{\rm 4} \\
\textsuperscript{\rm 1}College of Electronics and Information Engineering, Sichuan University, China \\
\textsuperscript{\rm 2}School of Computer Science, The University of Adelaide, Australia \\
\textsuperscript{\rm 3}University of Electronic Science and Technology of China, China\  \  \textsuperscript{\rm 4}Tencent \\
gouyanjie@stu.scu.edu.cn, yinjie@scu.edu.cn, lingqiao.liu@adelaide.edu.au\\ daiyongya@yahoo.com, lineshen@tencent.com}

\begin{document}
\maketitle
\begin{abstract}
Incorporating knowledge bases (KB) into end-to-end task-oriented dialogue systems is challenging, since it requires to properly represent the entity of KB, which is associated with \textit{its KB context} and \textit{dialogue context}.
The existing works represent the entity with only perceiving \textit{\textbf{a part of}} its KB context, which can lead to the less effective representation due to the information loss, and adversely favor KB reasoning and response generation.
To tackle this issue, we explore to \textit{\textbf{fully contextualize}} the entity representation by dynamically perceiving \textit{all the relevant entities} and \textit{dialogue history}.
To achieve this, we propose a COntext-aware Memory Enhanced Transformer framework (COMET), 
which treats the KB as a sequence and leverages a novel \textit{\textbf{Memory Mask}} to enforce the entity to only focus on its relevant entities and dialogue history, while avoiding the distraction from the irrelevant entities.
Through extensive experiments, we show that our COMET framework can achieve superior performance over the state of the arts.
\end{abstract}

\section{Introduction}

Task-oriented dialogue systems aim to achieve specific goals such as hotel booking and restaurant reservation. The traditional pipelines \cite{young-2017-pomdp,wen-etal-2017-network} consist of natural language understanding, dialogue management, and natural language generation modules. However, designing these modules often requires additional annotations such as dialogue states. To simplify this procedure, the end-to-end dialogue systems \cite{eric-manning-2017-copy} are proposed to incorporate the KB (normally relational databases) into the learning framework, where the KB and dialogue history can be directly modeled for response generation, without the explicit dialogue state or dialogue action.

\begin{table}
\begin{center}
\resizebox{1\columnwidth}{11mm}{%
\begin{tabular}{|l|l|l|l|l|}
\hline 
{\scriptsize{}Poi} & {\scriptsize{}Poi type} & {\scriptsize{}Traffic} & {\scriptsize{}Address} & {\scriptsize{}Distance}\tabularnewline
\hline 
\hline 
{\scriptsize{}Stanford Express Care} & {\scriptsize{}hospital} & {\scriptsize{}moderate} & {\scriptsize{}214 El Camino Real} & {\scriptsize{}2 miles}\tabularnewline
\hline 
\textcolor{blue}{\scriptsize{}Tom's house} & \textcolor{blue}{\scriptsize{}friend's house} & \textcolor{red}{\scriptsize{}no} & \textcolor{blue}{\scriptsize{}580 Van Ness Ave} & \textcolor{blue}{\scriptsize{}6 miles}\tabularnewline
\hline 
{\scriptsize{}Philz} & {\scriptsize{}coffee or tea place} & {\scriptsize{}no} & {\scriptsize{}583 Alester Ave} & {\scriptsize{}4 miles}\tabularnewline
\hline 
{\scriptsize{}5672 Barringer Street} & {\scriptsize{}certain address} & {\scriptsize{}no} & {\scriptsize{}5672 Barringer Street} & {\scriptsize{}2 miles}\tabularnewline
\hline
\end{tabular}}

\resizebox{1\columnwidth}{10mm}{%
\begin{tabular}{l|l}
\hline 
{\scriptsize{}User} & {\scriptsize{}Where does my friend live ?}\tabularnewline
\hline 
{\scriptsize{}System} & {\scriptsize{}\textcolor{teal}{Tom's house} is \textcolor{teal}{6 miles} away at \textcolor{teal}{580 Van Ness Ave} .}\tabularnewline
\hline 
{\scriptsize{}User} & {\scriptsize{}Is that the fastest route ?}\tabularnewline
\hline 
{\scriptsize{}System} & {\scriptsize{}I'll send the route with \textcolor{red}{no traffic} on your screen , drive carefully !}\tabularnewline
\hline 
\end{tabular}}

\caption{An example in SMD dataset \cite{eric-etal-2017-key}.
The top is the entities in KB and the bottom
is a two-turn dialogue between the user and system.}
\label{tab::example}
\end{center}
\end{table}

An example of the end-to-end dialogue systems is shown in Tab. \ref{tab::example}. When generating the second response about the ``\textit{traffic info}'':
(1) the targeted entity ``\textit{\textcolor{red}{no traffic}}'' is associated with its same-row entities (KB context) like ``\textit{\textcolor{blue}{Tome's house}}'',  ``\textit{\textcolor{blue}{friend's house}}'' and ``\textit{\textcolor{blue}{6 miles}}''. These entities can help with enriching the information of its representation and modeling the structure of KB. 
(2) Also, the entity is related to the dialogue history (dialogue context), which provides clues about the goal-related row (like ``\textit{\textcolor{teal}{Tom's house}}'' and ``\textit{\textcolor{teal}{580 Van Ness Ave}}'' in the first response). These clues can be leveraged to further enhance the corresponding representations and activate the targeted row, which benefits the retrieval of ``\textit{\textcolor{red}{no traffic}}''. Therefore, how to \textit{\textbf{fully contextualize}} the entity with its KB and dialogue contexts, is the key point of end-to-end dialogue systems \cite{madotto-etal-2018-mem2seq,wu2018globaltolocal,qin-etal-2020-dynamic}, where the full-context enhanced entity representation can make the reasoning over KB and the response generation much easier.

\begin{figure}[tb]
\begin{center}
\subfigure[Triplet rep.] {
 \label{fig::triplet-rep}     
\includegraphics[width=0.45\columnwidth]{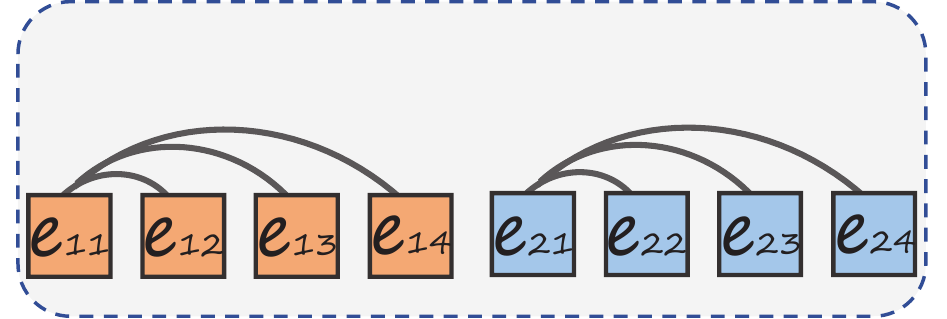}  
}     
\subfigure[Row-entity rep.] { 
\label{fig::row-ent-rep}     
\includegraphics[width=0.45\columnwidth]{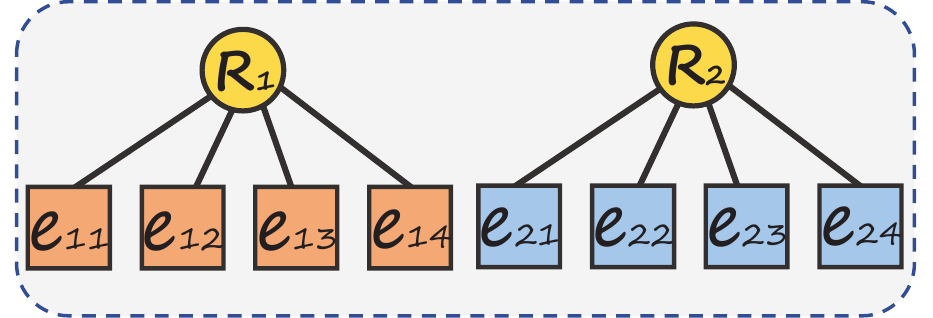}     
}    
\subfigure[Graph rep.] { 
\label{fig::graph-rep}     
\includegraphics[width=0.45\columnwidth]{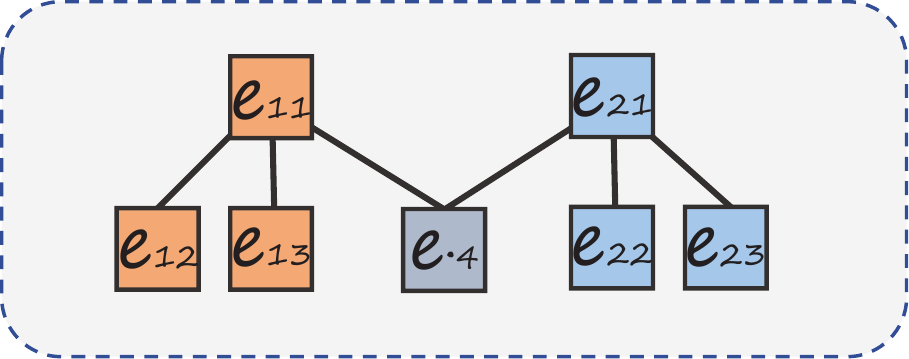}     
}   
\subfigure[Ours.] { 
\label{fig::our-rep}     
\includegraphics[width=0.45\columnwidth]{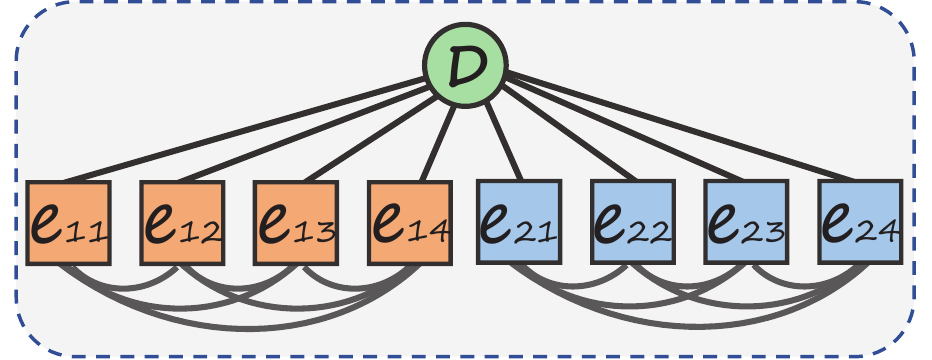}     
}
\end{center}
\caption{Four ways to represent the KB, where $e_{i,j}$ means the entity representation for the $j$-th entity of the $i$-th row; $R_i$ means the row representation of the $i$-th row; $e_{\cdot,j}$ means the entities shared between different row, like ``\textit{no traffic}'' in Tab. \ref{tab::example}; $D$ means the dialogue context. \textit{Note that the existing three representations (a-c) only consider parts of the KB context and ignore the dialogue context, whereas our method (d) can \textbf{fully contextualize} the entity with both of them.}
}     
\label{fig::diff-kb-rep-comp}     
\end{figure}

\textit{However, the existing works can only contextualize the entity with perceiving parts of its KB context and ignoring the dialogue context}: (1) \cite{madotto-etal-2018-mem2seq, wu2018globaltolocal, qin-etal-2020-dynamic} represent an entity as a triplet (cf. Fig. \ref{fig::triplet-rep}), i.e., (Subject, Relation, Object). However, breaking one row into several triplets can only model the relation between two entities, whereas the information from other same-row entities and dialogue history are ignored. (2) \cite{gangi-reddy-etal-2019-multi, qin-etal-2019-entity} represent KB in a hierarchical way, i.e., the row and entity-level representation (cf. Fig. \ref{fig::row-ent-rep}). This representation can only partially eliminate this issue at the row level. However, at the entity level, the entity can only perceive the information of itself, which is isolated with other KB and dialogue contexts. (3) \cite{yang-etal-2020-graphdialog} converts KB to a graph (cf. Fig. \ref{fig::graph-rep}). However, they fails to answer what is the optimal graph structure for KB. That indicates their graph structure may need manual design\footnote{For instance, on the SMD dataset, they only activate the edges between the primary key (``\textit{poi}'') and other keys(e.g., ``\textit{address}'') in the Navigation domain, but assign a fully-connected graph to the Schedule domain.}. Also, the dialogue context is not encoded into the entity representation, which can also lead to the suboptimal entity representation. 
To sum up, these existing methods \textit{\textbf{can not}} fully contextualize the entity, which leads to vulnerable KB reasoning and response generation.

\begin{figure*}[htb]
\begin{center}
\includegraphics[width=0.95\textwidth]{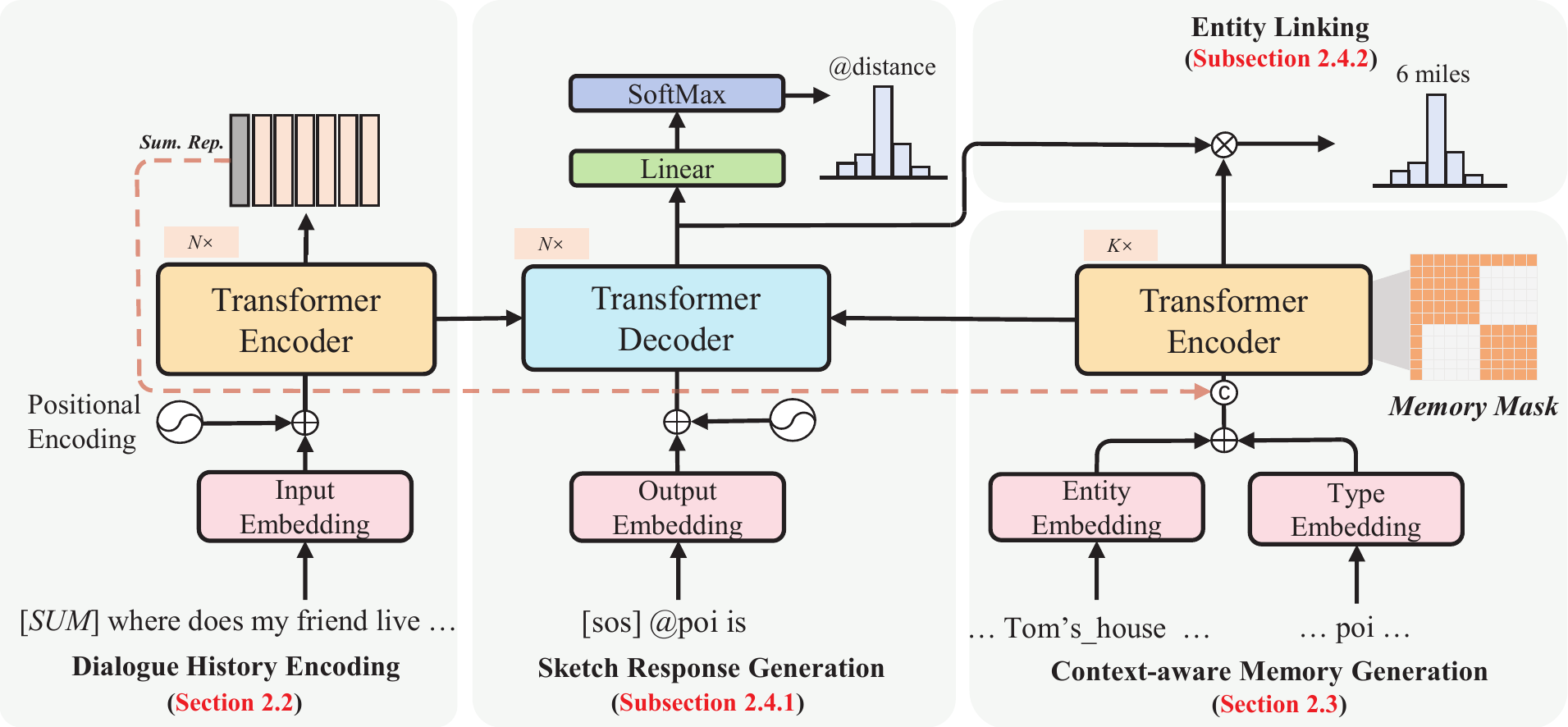}
\caption{Overview of \textbf{COMET}. The \textcolor{gray}{gray block} in the top left means Sum. Rep of dialogue history, \textit{which is used as the input for the Memory Generation.} \copyright \ means concatenation. The detailed construction of the Memory Mask can be found in Fig. \ref{fig::mask_construct}.
}
\label{fig::model_archi}
\end{center}
\end{figure*}

In this work, we propose COntext-aware Memory Enhanced Transformer (\textbf{COMET}), \textit{which provides a unified solution to \textbf{fully contextualize} the entity with the awareness of both the KB and dialogue contexts} (shown in Fig. \ref{fig::our-rep}). 
The key idea of COMET is that: a \textbf{\textbf{Memory-Masked}} Encoder is used to encode the entity sequence of KB, along with the information of dialogue history. The designed Memory Mask is utilized to ensure the entity can only interact with its same-row entities and the information in dialogue history, whereas the distractions from other rows are prohibited.

More specifically, (1) \textit{for the KB context}, we represent the entities in the same row as a sequence. Then, a Transformer Encoder \cite{vaswani-2017-attention} is leveraged to encode them, where the same-row entities can interact with each other. Furthermore, to retain the structure of KB and avoid the distractions from the entities in different rows, we design a \textit{\textbf{Memory Mask}} (shown in Fig. \ref{fig::mask_construct}) and incorporate it into the encoder, which only allows the interactions between the same-row entities.
(2) \textit{For the dialogue context}, we create a Summary Representation (\textit{Sum. Rep}) to summarize the dialogue history, which is input into the encoder to interact with the entity representations (gray block in Fig. \ref{fig::model_archi}). We also utilize the Memory Mask to make the \textit{Sum. Rep} overlook all of the entities for better entity representations, which will serve as the context-aware memory for further response generation.

By doing so, we essentially extend the entity of KB to $(\mathcal{N}+1)$-tuple representation, where $\mathcal{N}$ is the number of entities in one row and ``1'' is for the Sum. Rep of the dialogue history.
By leveraging the KB and dialogue contexts, our method can effectively model the information existing in KB and activate the goal-related entities, which benefits the entity retrieval and response generation.
Please note that the function of fully contextualizing entity is unified by the designed Memory Mask scheme, which is the key of our work.

We conduct extensive experiments on two public benchmarks, i.e., SMD \cite{eric-etal-2017-key, madotto-etal-2018-mem2seq} and Multi-WOZ 2.1 \cite{budzianowski-etal-2018-multiwoz, yang-etal-2020-graphdialog}. 
The experimental results demonstrate significant performance gains over the state of the arts. It validates that contextualizing KB with Transformer benefits entity retrieval and response generation.

In summary, our contributions are as follows:
\begin{itemize}
    \item To the best of our knowledge, we are the first to fully contextualize the entity representation with both the KB and dialogue contexts, for end-to-end task-oriented dialogue systems.
    \item We propose Context-aware Memory Enhanced Transformer, which incorporates a designed Memory Mask to represent entity with awareness of both the relevant entities and dialogue history.
    \item Extensive experiments demonstrate that our method gives a state-of-the-art performance.
\end{itemize}

\section{Methodology}
\label{sec:methodology}

In this section, we first introduce the general workflow for this task. Then, we elaborate on each part of COMET, i.e., the Dialogue History Encoder, Context-aware Memory Generation, and Response Generation Decoder (as depicted in Fig. \ref{fig::model_archi}). Finally, the objective function will be introduced.

\subsection{General Workflow}
\label{ssec:workflow}

Given a dialogue history with $k$ turns, which is denoted as $\mathcal{H} = \{u_1, s_1, u_2, s_2, ..., u_k\}$ ($u_i$ and $s_i$ denote the $i$-th turn utterances between the user and the system), the goal of dialogue systems is to generate the $k$-th system response $s_k$ with an external KB $\mathcal{B}=\{[b_{11}, ..., b_{1c}], ..., [b_{r1}, ..., b_{rc}]\}$, which has $r$ rows and $c$ columns. Formally, the procedure mentioned above is defined as: 
\begin{align}
    p(s_k|\mathcal{H}, \mathcal{B}) = \prod_{i=1}^{n}p(s_{k,t}|s_{k,1}, ..., s_{k,t-1}, \mathcal{H}, \mathcal{B}), \nonumber
\end{align}
where we first derive the dialogue history representation (Section \ref{ssec:encoder}) and generate the Context-aware Memory, a.k.a., contextualized entity representation (Section \ref{ssec:memory}), where these two parts will be used to generate the response $s_k$ (Section \ref{ssec:decoder}).

\subsection{Dialogue History Encoder}
\label{ssec:encoder}

We first transform $\mathcal{H}$ into the word-by-word form with a special token [\textit{SUM}]: $\mathcal{\hat{H}} = \{x_1, x_2, ..., x_n\},\ x_1=\ $[\textit{SUM}], which is used to globally aggregate information from $\mathcal{H}$. 

Then, the sequence $\mathcal{\hat{H}}$ is encoded by a standard Transformer Encoder and generate the dialogue history representation $H_{N}^{enc}$, where $H_{N,1}^{enc}$ is denoted as the Summary Representation (\textit{Sum. Rep}) of the dialogue history.\footnote{This module is as same as the standard Transformer Encoder, please refer to \cite{vaswani-2017-attention} for more details.}
\textit{It will be used to make the memory aware of the dialogue context.}

\subsection{Context-aware Memory Generation}
\label{ssec:memory}

In this subsection, we describe how to ``\textbf{\textit{fully contextualize KB}}”. That is, the  \textit{Memory Mask} is leveraged to ensure the entities of KB with the awareness of all of its related entities and dialogue history, which is the key contribution of our method.

\subsubsection{Memory Generation}
\label{sssec::rep_gen}

Different from existing works which fail to contextualize all the useful context information for the entity representation, we treat KB as a sequence, along with \textit{Sum. Rep}. Then, a Transformer Encoder with the Memory Mask is utilized to model it, which can dynamically generate the entity representation with the awareness of its all favorable contexts, i.e., the same-row entities and dialogue history, while blocking the distraction from the irrelevant entities. The procedure of memory generation is as follows.

Firstly, the entities in the KB $\mathcal{B}$ is flatten as a memory sequence, i.e., $\mathcal{M} = [b_{11}, ..., b_{1c}, ..., b_{r1}, ..., b_{rc}] = [m_1, m_2, ..., m_{|\mathcal{M}|}]$, where the memory entity $m_i$ means an entity of KB in the $k$-th row. By doing so, the Memory-Masked Transformer Encoder can interact the same-row entities with each other while retaining the structure information of KB.\footnote{When the memory sequence is long, some existing methods like the linear attention \cite{kitaev2020reformer} can be used to tackle the issue of $\mathcal{O}(N^2)$ complexity of Self Attention.}

Then, $\mathcal{M}$ will be transformed into the entity embeddings, i.e., $E = [e_1^m, ..., e_{|\mathcal{M}|}^m]$, where $e_i^m$ corresponds to $m_i$ in $\mathcal{M}$ and it is the sum of the word embedding $u_i$ and the type embedding $t_i$, i.e., $e_i^m = u_i + t_i$. Note that, the entity types are the corresponding column names, e.g., ``\textit{poi\_type}'' in Table \ref{tab::example}. For the entities which have more than one token, we simply treat them as one word, e.g., ``\textit{Stanford Exp}'' $\rightarrow$ ``\textit{Stanford\_Exp}''.

Next, the entity embeddings are concatenated with the \textit{Sum. Rep} from the Dialogue History Encoder, i.e. $E_0 = [H_{N,1}^{enc}; E]$. The purpose of introducing $H_{N,1}^{enc}$ is that it passes the information from the dialogue history and further enhances the entity representation with the dialogue context.

Finally, $E_0$ and the Memory Mask $M^{mem}$ are used as the input of the Transformer Encoder ($tf\_enc(\cdot)$) to generate the context-aware memory (a.k.a, contextualized entity representation):
\begin{gather}
    E_l =tf\_enc(E_{l-1}, M^{mem}), l \in [1, K], \nonumber
\end{gather}
where $K$ is the total number of Transformer Encoder layers. $E_K \in \mathbb{R}^{(|\mathcal{M}|+1) \times d_m}$  is the generated memory, which is queried when generating the response for entity retrieval.

\subsubsection{Memory Mask Construction}
\label{sssec::mem_mask}

To highlight, we design a special Memory Mask scheme to take ALL the contexts grounded by the entity into account, where the Memory Mask ensures that the entity can only attend to its context part, which is the key contribution of this work.
This is in contrast to the standard Transformer Encoder, where each entity can attend to all of the other entities. The rationale of our design is that by doing so, we can avoid the noisy distraction of the non-context part.

\begin{figure}[htb]
\begin{center}
\includegraphics[width=0.8\columnwidth]{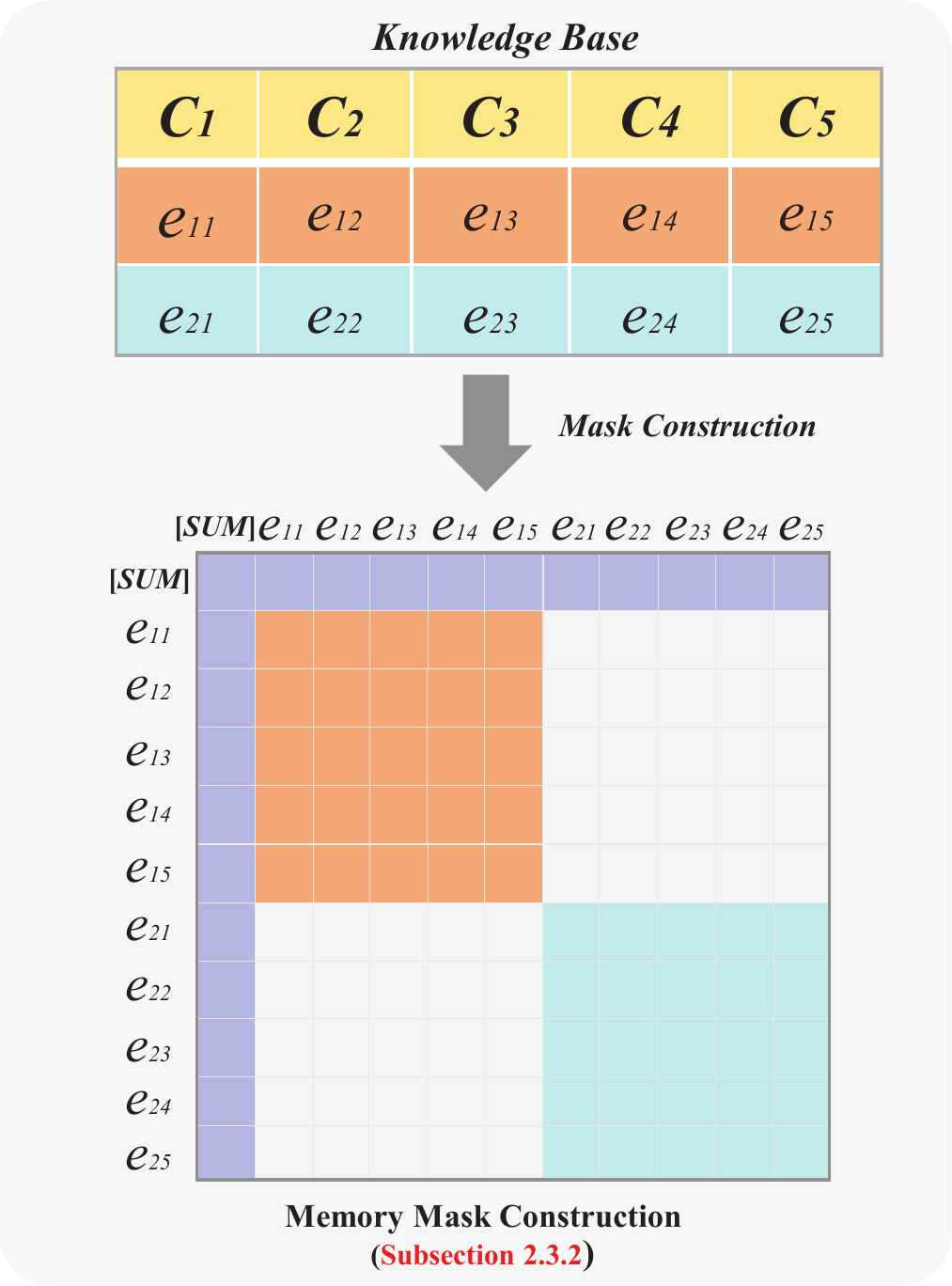}
\caption{The Construction of Memory Mask. $C_i$ means the column name (e.g., ``\textit{Poi}''). $e_{ij}$ means the $j$-th entity of $i$-th row. {[}\textit{SUM}{]} means the Sum. Rep. Only two rows of KB are shown for simplicity.
}
\label{fig::mask_construct}
\end{center}
\end{figure}

Formally, $M^{mem} \in \mathbb{R}^{(|\mathcal{M}|+1) \times (|\mathcal{M}|+1)}$ is defined as:
\begin{equation}
M^{mem}_{i,j}=\left\{
\begin{aligned}
&1, &\ if\  \mathcal{M}_{i-1}, \mathcal{M}_{j-1} \in b_k, \\
&1, &\ if\  i\ or\  j = 1, \\
&-\infty, &\ else. \nonumber
\end{aligned}
\right.
\end{equation}
A detailed illustration of the Memory Mask construction is shown in Fig. \ref{fig::mask_construct}. 
With this designed Memory Mask, a masked attention mechanism is leveraged to make the entity only attend the entities within the same row and the \textit{Sum. Rep}.

\subsection{Response Generation Decoder}
\label{ssec:decoder}

Given the dialogue history representation $H_N^{enc}$ and generated memory $E_K$, the decoder will use them to generate the response for a specific query. In COMET, we use a modified Transformer Decoder, which has two cross attention modules to model the information in $H_N^{enc}$ and $E_K$, respectively. Then, a gate mechanism is leveraged to adaptively fuse $H_N^{enc}$ and $E_K$ for the decoder, where the response generation is tightly anchored by them.

Following \cite{wu2018globaltolocal,qin-etal-2020-dynamic, yang-etal-2020-graphdialog}, we first generate a sketch response that replaces the exact slot values with sketch tags.\footnote{For instance, ``Tom’s house is 6 miles away at 580 Van Ness Ave ." $\rightarrow$ ``@poi is @distance away at @address.".} Then, the decoder links the entities in the memory to their corresponding slots.

\subsubsection{Sketch Response Generation}
\label{sssec:sketch_reponse}

For the $k$-th turn generating sketch response $\mathcal{Y} = [y_1, ... y_{t-1}]$, it is converted to the word representation $H_0^{dec} = [w_1^d, ..., w_{t-1}^d]$.
$w_i^d = v_i + p_i$, where $v_i$ and $p_i$ means the word embedding and absolute position embedding of $i$-th token in $\mathcal{Y}$.

Afterward, $N$-stacked decoder layers are applied to decode the next token with the inputs of $H_0^{dec}$, $E_K$ and $H_N^{enc}$. The process in one decoder layer can be expressed as:
\begin{align}
    H_{l}^{d-d} &= MHA(H_{l-1}^{dec}, H_{l-1}^{dec}, H_{l-1}^{dec}, M^{dec}), \nonumber\\
    H_{l}^{d-e} &= MHA(H_{l}^{d-d}, H_N^{enc}, H_N^{enc}),  \nonumber\\
    H_{l}^{d-m} &= MHA(H_{l}^{d-d}, E_K, E_K),  \nonumber\\
    g &= sigmoid(FC(H_{l}^{d-m})),  \nonumber \\
    H_{l}^{agg} &= g \odot H_{l}^{d-e} + (1 - g) \odot H_{l}^{d-m},  \nonumber\\
    H_{l}^{dec} &= FFN(H_{l}^{agg}), \ l \in [1, N], \nonumber
\end{align}
where the input $\{Q,K,V,M\}$ of the Multi-Head Attention $MHA(Q, K, V, M)$ means the query, key, value, and optional attention mask. $FFN(\cdot)$ means the Feed-Forward Networks. $M^{dec}$ is the decoder mask, so as to make the decoded word can only attend to the previous words.
$FC(\cdot)$ is a fully-connected layer to generate the gating signals, which maps a $d_m$-dimension feature to a scalar. $N$ is the number of the total decoder layers.

After obtaining the final $H_{N}^{dec}$, the posterior distribution for the $t$-th token, $p_t^{v} \in \mathbb{R} ^{|V|}$ ($|V|$ denotes the vocabulary size), is calculated by:
\begin{gather}
    p_t^{v} = softmax(H_{N,t-1}^{dec}W_v + b_v). \nonumber
\end{gather}

\subsubsection{Entity Linking}
\label{ssec:linking}

After the sketch response generation, we replace the sketch tags with the entities in the context-aware memory. We denote the representation from the decoder at the $t$-th time step, i.e., the $t$-th token, as $H_{N,t}^{dec}$, and represent the time steps that need to replace sketch tags with entities as $\mathcal{T}$. The probability distribution over all possible linked entities can then be calculated by 
\begin{gather}
    p^s_t = softmax(H_{N,t}^{dec}E_K^T),~~\forall t \in \mathcal{T} \nonumber
\end{gather}
where $E_K$ means the final generated memory.

\subsection{Objective Function}

For the training process of COMET, we use the the cross-entropy loss to supervise the response generation and entity linking\footnote{The label construction procedure of the entity linking module can be found in Appendix A.1.}.

Moreover, we propose an additional regularization term to further regularize $p^s_t$. The regularization is based on the prior knowledge that for a given response, only a small subset of entities should be linked. Formally, we construct 
the following entity linking probability matrix  $\mathbf{P}^{s} = [p^s_{t_1}, p^s_{t_2}, ..., p^s_{t_{|\mathcal{T}|}}]$ and minimize its $L_{2,1}$-norm \cite{nie2010efficient}:
\begin{gather}
    L_r = \sum_{i=1}^{|\mathcal{M}|}\sqrt{\sum_{t \in \mathcal{T}}(p^s_{t, i})^2} \ , \nonumber
\end{gather}
where $p^s_{t, i}$ denotes the $i$-th dimension of $p^s_{t}$. This regularization term can encourage the network to select a small subset of entities to generate the response. The same idea has been investigated in \cite{nie2010efficient} for multi-class feature selection.

Finally, COMET is trained by jointly minimizing the combination of the above three losses.

\begin{table*}[htb]
\centering
\resizebox{2.0\columnwidth}{21mm}{%
\begin{tabular}{c|c|c|c|c|c|c|c|c|c|c|c}
\hline 
 & \multicolumn{5}{c|}{SMD} & \multicolumn{6}{|c}{Multi-WOZ2.1}\tabularnewline
\hline 
\hline 
Model & BLEU & F1 & F1-Sch. & F1-Wea. & F1-Nav. & BLEU & F1 & F1-Res. & F1-Att. & F1-Hot. & F1-Tra.  \tabularnewline
\hline 
Mem2Seq & 12.6 & 33.4 & 49.3 & 32.8 & 20.0 & 4.1 & 3.2 &  2.9 & 2.1  & 4.5 & 1.5 \tabularnewline
\hline 
KB-Transformer & 13.9 & 37.1 & 51.2 & 48.2 & 23.3 & - & - & - & - & - & - \tabularnewline
\hline 
KB-Retriever & 13.9 & 53.7 & 55.6 & 52.2 & 54.5 & - & - & - & - & - & - \tabularnewline
\hline 
GLMP & 13.9 & 60.7 & 72.9 & 56.5 & 54.6 & 4.3  & 6.7 & 11.4 & 9.4  & 3.9  & 3.5\tabularnewline
\hline 
DF-Net & \underline{14.4} & \underline{62.7} & \underline{73.1} & \underline{57.6} & \textbf{57.9} & - & -  & -  & -  & -  & - \tabularnewline
\hline
GraphDialog & 13.7 & 60.7 & 72.8 & 55.2 & 54.2 & \underline{6.2} & \underline{11.3} & \underline{16.0} & \underline{14.1} & \underline{10.8} & \underline{4.4} \tabularnewline
\hline 
COMET (\textit{Ours}) & \textbf{17.3} & \textbf{63.6} & \textbf{77.6} & \textbf{58.3} & \underline{56.0} &  \textbf{8.3}  & \textbf{18.6}  & \textbf{27.5}  & \textbf{17.9}  & \textbf{15.2} & \textbf{9.8} \tabularnewline
\hline 
\end{tabular}{\scriptsize\par}}
\caption{BLEU and Entity F1 comparison of COMET with other counterparts. The best results are in \textbf{bold font} and the second-best results are \underline{underlined}. 
The results on the SMD and Multi-WOZ 2.1 datasets are adopted from \cite{qin-etal-2020-dynamic} and \cite{yang-etal-2020-graphdialog}, respectively.
}
\label{tab:results}
\end{table*}

\section{Experiments}
\label{sec:experiments}

\subsection{Datasets}
\label{ssec:datasets}
Two public multi-turn task-oriented dialogue datasets are used to evaluate our model, i.e., SMD\footnote{\url{https://github.com/jasonwu0731/GLMP/tree/master/data/KVR}} \cite{eric-etal-2017-key} and Multi-WOZ 2.1\footnote{\url{https://github.com/shiquanyang/GraphDialog/tree/master/data/MULTIWOZ2.1}} \cite{budzianowski-etal-2018-multiwoz}. \textit{Note that, for Multi-WOZ 2.1, to accommodate end-to-end settings, we use the revised version released by \cite{yang-etal-2020-graphdialog}, which equips the corresponding KB to every dialogue.} We follow the same partition as \cite{madotto-etal-2018-mem2seq} on SMD and \cite{yang-etal-2020-graphdialog} on Multi-WOZ 2.1. 

\subsection{Experimental Settings}
The dimension of embeddings and hidden vectors are all set to 512. The number of layers ($N$) in Dialogue History Encoder and Response Generation Decoder is set to 6. The number of layers for Context-aware Memory Generation ($K$) is set to 3. The number of heads in each part of COMET is set to 8. A greedy strategy is used without beam-search during decoding. The Adam optimizer \cite{kingma2014adam} is used to train our model from scratch with a learning rate of $1e^{-4}$. More details about the hyper-parameter settings can be found in Appendix A.2.

\subsection{Baselines}
We compare COMET with the following methods: 
\begin{itemize}
    \item \textbf{Mem2Seq} (\textit{Triplet}) \cite{madotto-etal-2018-mem2seq}: Mem2Seq incorporates the multi-hop attention mechanism in memory networks into the pointer networks.
    \item \textbf{KB-Transformer} (\textit{Triplet}) \cite{h-2019-kbtransformer}: KB-Transformer combines a Multi-Head Key-Value memory network with Transformer.
    \item \textbf{KB-Retriever} (\textit{Row-entity}) \cite{qin-etal-2019-entity}: KB-retriever improves the entity-consistency by first selecting the target row and then picking the relevant column in this row.
    \item \textbf{GLMP} (\textit{Triplet}) \cite{wu2018globaltolocal}: GLMP uses a global memory encoder and a local memory decoder to incorporate the external knowledge into the learning framework.
    \item \textbf{DF-Net} (\textit{Triplet}) \cite{qin-etal-2020-dynamic}: DF-Net applies a dynamic fusion mechanism to transfer knowledge in different domains.
    \item \textbf{GraphDialog} (\textit{Graph}) \cite{yang-etal-2020-graphdialog}: GraphDialog exploits  the  graph  structural information in KB and in the dependency parsing tree of the dialogue.
\end{itemize}

\subsection{Results}
Following the existing works \cite{qin-etal-2020-dynamic, yang-etal-2020-graphdialog}, we use the \textit{BLEU} and \textit{Entity F1}
metrics to evaluate model performance. The results are shown in Tab. \ref{tab:results}.

It is observed that: COMET achieves the best performance over both datasets, which indicates that our COMET framework can better leverage the information in the dialogue history and external KB, to generate more fluent responses with more accurate linked entities. Specifically, for the \textit{BLEU} score, it outperforms the previous methods by 2.9\% on the SMD dataset and 2.1\% on the Multi-WOZ 2.1 dataset, at least. Also, COMET achieves the highest \textit{Entity F1} score on both datasets. That is, the improvements of 0.9\% and 7.3\% are attained on the SMD and Multi-WOZ 2.1 datasets, respectively. In each domain of the two datasets, improvement or competitive performance can be clearly observed. The results indicate the superior of our COMET framework.

To highlight, KB-Transformer \cite{h-2019-kbtransformer} also leverages Transformer, but our COMET outperforms it by a large margin. On the SMD dataset, the \textit{BLEU} score of COMET is higher than that of KB-Transformer by 3.4\%. The improvement introduced by COMET on \textit{Entity F1} score is as significant as 26.5\%. This shows naively introducing Transformer to the end-to-end dialogue system will not necessarily lead to higher performance. A careful design of the whole dialogue system, such as our proposed one, plays a vital role.

\subsection{Ablation Study}
\label{ssec::ablation}
In this subsection, we first investigate the effects of the different components, i.e., the Memory Mask, Sum. Rep, gate mechanism, and $L_{2,1}$-norm regularization (Tab. \ref{tab:ablation_2}). 
Then, we design careful experiments to further demonstrate the effect of the Memory Mask, which is the key contribution of this work:
(1) we replace the context-aware memory of COMET with the existing three representations of KB, (i.e., triplet, row-entity, and graph) to show the superior of the fully contextualized entity (Tab. \ref{tab:transformer+x}).
(2) We also replace our Memory Mask with the full attention layer by layer, which further shows the importance of our Memory Mask (Tab. \ref{tab:comb_mem}).
Our ablation studies are based on the SMD dataset.

\begin{table}[htb]
\centering

\resizebox{0.9\columnwidth}{16mm}{%
\begin{tabular}{l|c||c|c}
\hline 
Model & BLEU  & Entity F1 & $\Delta$  \tabularnewline
\hline 
\hline 
COMET & 17.3 & 63.6 & - \tabularnewline
\hline 
w/o Memory Mask & 15.4  & 49.6 & 14.0  \tabularnewline
\hline 
w/o Sum. Rep & 17.0 & 61.4 & 2.2 \tabularnewline
\hline
only use $H_N^{enc}$ (gate) & 17.2  & 61.1 & 2.5 \tabularnewline
\hline 
only use $E_K$ (gate) & 17.1  & 61.4 & 2.2 \tabularnewline
\hline 
w/o $L_{2,1}$-norm & 17.4  & 62.3 & 1.3 \tabularnewline
\hline 
\end{tabular}{\scriptsize{}}}
\caption{The effects of different components.}
\label{tab:ablation_2}
\end{table}

The effects of the key components in the COMET framework are reported in Tab. \ref{tab:ablation_2}. As observed, removing any key component of the COMET, both the \textit{BLEU} and \textit{Entity F1} metrics degrade to some extend. More specifically: 
(1) If the Memory Mask is removed, the \textit{Entity F1} score drops to 49.6. This significant discrepancy demonstrates the importance of restricting self-attention as our designed Memory Mask did.
(2) For the variant without the Sum. Rep, the \textit{Entity F1} score drops to 61.4. That indicates the effectiveness of contextualizing the KB with the dialogue history, which can further boost the performance.
(3) We also remove the gate and only use the information from the dialogue history ($H_N^{enc}$) or memory ($E_K$). We can see that the former case can only achieve 61.1 while the latter case achieves 61.4 of the \textit{Entity F1} score. It is obvious that using the gate mechanism to fuse both information sources is helpful for the entity linking.  
(4) When removing the $L_{2,1}$-norm, the performance also drops to 62.3, which means regularizing the entity-linking distribution can further benefit the performance.

\begin{table}[htb]
\centering
\resizebox{0.95\columnwidth}{12mm}{%
\begin{tabular}{l|c|c|c|c|c}
\hline 
Model & BLEU & F1 & F1-Sch. & F1-Wea. & F1-Nav. \tabularnewline
\hline 
\hline 
Context-aware memory & \textbf{17.3} & \textbf{63.6} & \textbf{77.6} & \textbf{58.3} & \textbf{56.0} \tabularnewline
\hline 
Only KB context & \underline{17.0} & \underline{61.4} & \underline{75.5}  & \underline{55.2}  & \underline{54.4}   \tabularnewline
\hline
Triplet & 14.9 & 59.8 & 73.1 & 54.0 & 53.0 \tabularnewline
\hline 
Row\&Ent & 13.0 & 41.4 & 51.2 & 54.6 & 19.3 \tabularnewline
\hline 
Graph & 14.4 & 56.7 & 71.6 & 48.7 & 50.4 \tabularnewline
\hline 
\end{tabular}{\scriptsize\par}}
\caption{The performance of replacing the context-aware memory with Triplet, Row-Ent and Graph representations in COMET. Note that in the second row, we also report the result of a variant which only considers the KB context and ignores the dialogue context.}
\label{tab:transformer+x}
\end{table}

We also replace our context-aware memory with other ways of representing KB, while other parts of our framework keep unchanged\footnote{The implementation details are in Appendix A.3.}. The result is reported in Tab. \ref{tab:transformer+x}. It is observed that, After replacing our context-aware memory with the existing three representations of KB, the performance drops a lot in all the metrics, where the \textit{BLEU} score drops 2.4\% and the \textit{Entity F1} score drops 3.8\% at least. 
Besides, the result of the variant which only considers the KB context part (i.e., w/o Sum. Rep), is also reported, so as to further fairly compare with the aforementioned KB representations. The result shows that only considering the KB context, our method can still outperform other KB representations by 1.6\% of \textit{Entity F1} at least. That further indicates the fully contextualizing entity with its relevant entity and the dialogue history, can better represent the KB for dialogue systems.

\begin{table}[htb]
\centering
\resizebox{0.7\columnwidth}{12mm}{%
\begin{tabular}{l|c||c|c}
\hline 
Scheme & BLEU & Entity F1 & $\Delta$ \tabularnewline
\hline 
\hline 
MMM & 17.3 & 63.6 & - \tabularnewline
\hline 
MMF & 16.5 & 61.2 & 2.4 \tabularnewline
\hline 
MFF & 16.5 & 59.1 & 4.5 \tabularnewline
\hline 
FFF & 15.4 & 49.6 & 14.0 \tabularnewline
\hline 
\end{tabular}{\scriptsize\par}}
\caption{The performance of replacing the Memory Mask with the full attention. The meanings of the scheme names are that the Memory Mask (\textbf{M}) is replaced with the Full attention (\textbf{F}).}
\label{tab:comb_mem}
\end{table}

\begin{table*}[htbp]
\centering
\resizebox{2.\columnwidth}{56mm}{%
\begin{tabular}{l}
\hline 
Query\&Response Example\tabularnewline
\hline 
\hline 
Goal: {[}yoga\_activity, 11am, thursday, alex{]}\tabularnewline
Query: what time do i go to yoga and who is going with me ?\tabularnewline
Response: yoga is with {[}alex{]} at {[}11am{]}.\tabularnewline
GLMP: your yoga is on {[}\textbf{thursday}{]}$_\checkmark$ with {[}\textbf{alex}{]}$_\textbf{\checkmark}$.\tabularnewline
DF-Net: your yoga is on {[}\textbf{thursday}{]}$_\checkmark$ at {[}\textbf{5pm}{]}\textcolor{red}{$_\times$}.\tabularnewline
COMET: your yoga is on {[}\textbf{thursday}{]}$_\checkmark$ at {[}\textbf{11am}{]}$_\checkmark$ with {[}\textbf{alex}{]}$_\checkmark$. (\textit{More informative response.})\tabularnewline
\hline 
Goal: {[}stanford\_express\_care, 6\_miles, heavy\_traffic, hospital,
214\_el\_camino\_real{]}\tabularnewline
Query: car i need a hospital , please send me a quick route to the one
near me in the direction we are driving.\tabularnewline
Response: stanford\_express\_care is on the way.\tabularnewline
GLMP: stanford\_express\_care is {[}\textbf{2\_miles}{]}\textcolor{red}{$_\times$} away. (\textit{Wrong distance info.})\tabularnewline
DF-Net: stanford\_express\_care is {[}\textbf{4\_miles}{]}\textcolor{red}{$_\times$} away. (\textit{Wrong distance info.})\tabularnewline
COMET: stanford\_express\_care is {[}\textbf{6\_miles}{]}$_\checkmark$ away.\tabularnewline
\hline 
Goal: {[}mountain\_view, friday, clear\_skies, 30f, 50f{]}\tabularnewline
Query: what will be the highest\_temperature on friday in mountain\_view
?\tabularnewline
Response: on friday , the highest\_temperature in mountain\_view is predicted
to be {[}50f{]}.\tabularnewline
GLMP: it will be friday on mountain\_view in highest\_temperature.  \textcolor{red}{$\times$} (\textit{Fail to retrieve the temperature info.}) \tabularnewline
DF-Net: it will be highest\_temperature on friday in mountain\_view.  \textcolor{red}{$\times$} (\textit{Fail to retrieve the temperature info.}) \tabularnewline
COMET: on friday the temperature in mountain\_view will be low of {[}\textbf{30f}{]}$_\checkmark$
, and high of {[}\textbf{50f}{]}$_\checkmark$.\tabularnewline
\hline 
Goal: {[}teavana, 5\_miles, moderate\_traffic, coffee\_or\_tea\_place, 145\_amherst\_st{]}\tabularnewline
Query: get me directions to the nearest starbucks.\tabularnewline
Response: there is \textbf{NO} starbucks locally.\tabularnewline
GLMP: there is a starbucks {[}\textbf{1\_miles}{]}\textcolor{red}{$_\times$} away. (\textit{Unfaithful response.}) \tabularnewline
DF-Net: the nearest starbucks is teavana , it s {[}\textbf{1\_miles}{]}\textcolor{red}{$_\times$} away. (\textit{Not fluent and wrong entities retrieved.}) \tabularnewline
COMET: there is \textbf{NO} starbucks nearby , but {[}\textbf{teavana}{]}$_\checkmark$ is {[}\textbf{5\_miles}{]}$_\checkmark$ away would you like directions there?\tabularnewline
\hline
\end{tabular}{\scriptsize{}}}
\caption{Responses generated by our COMET, GLMP \cite{wu2018globaltolocal} and DF-Net \cite{qin-etal-2020-dynamic} from the SMD dataset. Goal means the row that the user is queried.
$\checkmark$ and \textcolor{red}{$\times$} mean the right or wrong entity linked.}
\label{tab::case_study}
\end{table*}

We also conduct the experiment which replaces the Memory Mask with the full attention, layer by layer. That is, the first (n-k) layers use the proposed Memory Mask (\textbf{M}) and the last k layers use the full attention (\textbf{F}).
As shown in Tab. \ref{tab:comb_mem}, the more full attention is added, the more performance of COMET drops in all of the metrics since the full attention introduces too much distraction from other rows. 
The result further indicates that the Memory Mask is indeed a better choice which takes the inductive bias of KB into account.

Note that we also explore other Memory Mask schemes, but these schemes can not further boost the performance, where the results are omitted due to the page limitation. For further improvement, more advanced techniques like Pre-trained Model \cite{devlin2018bert,radford2019language} may be needed to deeply understand the dialogue and KB context, which we leave for future work.

\subsection{Case Study}
To demonstrate the superiority of our method, several examples on the SMD test set, which are generated by our COMET and the existing state of the arts GLMP \cite{wu2018globaltolocal} and DF-Net \cite{qin-etal-2020-dynamic}, are given in Tab. \ref{tab::case_study}. 
As reported, compared with GLMP and DF-Net, COMET can generate more fluent, informative, and accurate responses. 

Specifically, 
in the first example, GLMP and DF-NET are lack of the necessary information ``\textit{11am}'' or provide the wrong entity ``\textit{5pm}''. But COMET can obtain all the correct entities, 
which is more informative. 
In the second example, our method can generated the response with the right ``\textit{distance}'' information but GLMP and DF-Net can not.
In the third example, GLMP and DF-Net can not even generate a fluent response, let alone the correct temperature information. But COMET can still perform well. 
The fourth example is more interesting: the user queries the information about ``\textit{starbucks}'' which does not exist in the current KB. GLMP and DF-Net both fail to faithfully respond, whereas COMET can better reason KB to generate the right response and even provide an alternative option.

\section{Related Work}
\label{sec:related_work}

Task-oriented dialogue systems can be mainly categorized into two parts: modularized \cite{williams2007partially,wen-etal-2017-network} and  end-to-end \cite{eric-manning-2017-copy}. 
For the end-to-end task-oriented dialogue systems, \cite{eric-manning-2017-copy} first explores the end-to-end method for the task-oriented dialogue systems. However, it can only link to the entities in the dialogue context and no KB is incorporated. 
To effectively incorporate the external KB, \cite{eric-etal-2017-key} proposes a key-value retrieval mechanism to sustain the grounded multi-domain discourse. 
\cite{madotto-etal-2018-mem2seq} augments the dialogue systems with end-to-end memory networks \cite{sukhbaatar-2015-memory}. 
\cite{wen-etal-2018-sequence} models a dialogue state as a fixed-size distributed representation and uses this representation to query KB. 
\cite{lei-etal-2018-sequicity} designs belief spans to track dialogue believes, allowing task-oriented dialogue systems to be modeled in a sequence-to-sequence way. 
\cite{gangi-reddy-etal-2019-multi} proposes a multi-level memory to better leverage the external KB.
\cite{wu2018globaltolocal} proposes a global-to-local memory pointer network to reduce the noise caused by KB.
\cite{lin-etal-2019-task} proposes Heterogeneous Memory Networks to handle the heterogeneous information from different sources.
\cite{qin-etal-2020-dynamic} proposes a dynamic fusion mechanism to transfer the knowledge among different domains. 
\cite{yang-etal-2020-graphdialog} exploits the graph structural information in KB and the dialogue.
Other works also explore how to combine the Pre-trained Model \cite{devlin2018bert,radford2019language} with the end-to-end task-oriented dialogue systems. \cite{madotto2020learning} directly embeds the KB into the parameters of GPT-2 \cite{radford2019language} via fine-tuning. 
\cite{madotto2020adapter} proposes a dialogue model that is built with a fixed pre-trained conversational model and multiple trainable light-weight adapters. 

We also notice that some existing works also combine Transformer with the memory component, e.g., \cite{ma2021streaming}. 
However, our method is distinguishable from them, since the existing works like \cite{ma2021streaming} simply inject the memory component into Transformer.
In contrast, inspired by the dynamic generation mechanism \cite{GOU2020105912}, the memory in COMET (i.e., the entity representation) is dynamically generated by fully contextualizing the KB and dialogue context via the Memory-masked Transformer.

\section{Conclusion}
\label{sec:conclusion}

In this work, we propose a novel COntext-aware Memory Enhanced Transformer (COMET) for the end-to-end task-oriented dialogue systems. By the designed Memory Mask scheme, COMET can fully contextualize the entity with all its KB and dialogue contexts, and generate the $(\mathcal{N}+1)$-tuple representations of the entities. The generated entity representations can further augment the framework and lead to better capabilities of response generation and entity linking. The extensive experiments demonstrate the effectiveness of our method.

\section*{Acknowledgements}
We would like to thank the anonymous reviewers for their valuable comments.

\bibliography{anthology,custom}
\bibliographystyle{acl_natbib}

\clearpage

\appendix

\section{Appendices}

\subsection{Label Construction of Entity Linking}
In practice, the datasets do not provide the golden linked entity. However, We could obtain a pseudo annotation by following \cite{qin-etal-2019-entity} to use a distant supervision method. Specifically, we match the entities in the golden response against the entities in the memory $\mathcal{M}$ and use the matching result as the golden entity. For entities like ``no\_traffic'', one may find matches in multiple rows. We resolve this ambiguity by choosing the entity from the row which has the most matches for all entities in the utterances.

\subsection{Hyper-parameter Settings}

\begin{table}[htb]
\centering
\resizebox{0.96\columnwidth}{25mm}{%
\begin{tabular}{l|c|c}
\hline 
Hyper-parameter & SMD  & Multi-WOZ 2.1 \tabularnewline
\hline 
\hline 
Batch Size & 32 & 16\tabularnewline
\hline 
Hidden Size & 512  & 512\tabularnewline
\hline 
Embedding Size & 512  & 512\tabularnewline
\hline 
\#Layer of Dialogue Enc. & 6  & 6\tabularnewline
\hline 
\#Layer of Response Dec. & 6  & 6\tabularnewline
\hline 
\#Layer for Memory & 3 & 3\tabularnewline
\hline 
\#Head & 8 & 8\tabularnewline
\hline 
Learning Rate & 0.0001 & 0.0001\tabularnewline
\hline 
KB Mask Prob. & 0.2 &  0.05 \tabularnewline
\hline 
Dropout Prob. & 0.1 &  0.1\tabularnewline
\hline 
\end{tabular}}
\caption{Hyper-parameters used in the two datasets.}
\label{tab:hyper-para}
\end{table}

We follow \cite{wu2018globaltolocal} to randomly mask a small number of entities into an unknown token to improve the generalization of our model. Besides, in the sketch generation and entity linking stages, we also use the label smoothing to regularize the model. The hyper-parameters such as dropout rate are tuned over the development set by grid search (\textit{Entity F1} for both datasets). The model is implemented in PyTorch. 
The hyper-parameters used in two datasets are shown in Tab. \ref{tab:hyper-para}.

\subsection{Implementation Details of Other KB Representations with Transformer}

To further compare the different methods of representing KB with our method, we also adopt the triplet, row-entity, and graph representation to replace our contextualized entity representation, where we keep the other parts of COMET unchanged. 

Specifically, for the triplet representation, we follow \cite{madotto-etal-2018-mem2seq, wu2018globaltolocal, qin-etal-2020-dynamic} to implement Transformer+Triplet, where the entity representation is the sum of the subject, relation, and object. Besides, the multi-hop reasoning \cite{sukhbaatar-2015-memory} is leveraged to further boost the performance. For the row-ent representation, we refer to \cite{gangi-reddy-etal-2019-multi, qin-etal-2019-entity} to implement Transformer+Row\&Ent, where Bag-of-word embedding and entity-type embedding are used for the row-level representation and entity-level representation. Besides, the row-level representation and entity-level representation are hierarchically queried, where the distribution of the entity-level embedding is used for the response generation. For the graph representation, we adopt the memory part of GraphDialog \cite{yang-etal-2020-graphdialog} to implement Transformer+Graph, where the entity embedding is further augmented by Graph Neural Networks \cite{velickovic2018graph}. 
Besides, the last hop of the triplet and graph representation, and the entity-level representation of Row\&Entity representation will be also used to adaptively fuse the information of KB in the Decoder of COMET.
More details can be found in the aforementioned papers.

\end{document}